\pdfoutput=1

\documentclass[11pt]{article}

\usepackage[]{acl}

\usepackage{times}
\usepackage{latexsym}

\usepackage[T1]{fontenc}

\usepackage[utf8]{inputenc}

\usepackage{microtype}

\usepackage{inconsolata}
\usepackage{natbib}

\usepackage[T1]{fontenc}    
\usepackage{hyperref}       
\usepackage{url}            
\usepackage{booktabs}       
\usepackage{amsfonts}       
\usepackage{nicefrac}       
\usepackage{xcolor}         

\usepackage{amsmath}
\usepackage{graphicx}
\usepackage{multirow}
\usepackage{pifont}
\usepackage{textcomp}
\title{Learning to Retrieve In-Context Examples for Large Language Models}

\author{Liang Wang \and Nan Yang \and Furu Wei \\
        Microsoft Research \\
        \{wangliang,nanya,fuwei\}@microsoft.com}

\begin{document}
\maketitle

\begin{abstract}
    Large language models (LLMs) have demonstrated their ability to learn in-context,
    allowing them to perform various tasks based on a few input-output examples.
    However, the effectiveness of in-context learning is heavily reliant on the quality of the selected examples.
    In this paper,
    we propose a novel framework to iteratively train dense retrievers that
    can identify high-quality in-context examples for LLMs.
    Our framework initially trains a reward model based on LLM feedback to evaluate the quality of candidate examples,
    followed by knowledge distillation to train a bi-encoder based dense retriever.
    Our experiments on a suite of $30$ tasks demonstrate that
    our framework significantly enhances in-context learning performance.
    Furthermore, we show the generalization ability of our framework to unseen tasks during training.
    An in-depth analysis reveals that our model improves performance by
    retrieving examples with similar patterns,
    and the gains are consistent across LLMs of varying sizes.
    The code and data are available at \url{https://github.com/microsoft/LMOps/tree/main/llm_retriever}.
\end{abstract}

\section{Introduction}
In-context learning (ICL) ~\citep{NEURIPS2020_1457c0d6} is an emerging learning paradigm that
allows LLMs to perform tasks with few-shot examples,
without requiring any updates to the model parameters.
This approach stands in stark contrast to traditional machine learning,
where models are typically trained on large datasets of labeled examples ~\citep{kenton2019bert}.
In-context learning offers a significant advantage in domains where labeled data is scarce or expensive to obtain,
as it greatly reduces the amount of required labeled data.

There are several challenges associated with understanding and enhancing the effectiveness of in-context learning.
One such challenge is that LLMs can be highly sensitive to the quality of
the in-context examples provided ~\citep{liu2022makes,min2022rethinking}.
If the examples are not representative of the target task,
then the model may not be able to learn effectively.
Empirical studies ~\citep{liu2022makes,luo2023dr} have demonstrated that
using BM25 algorithm or off-the-shelf sentence embeddings ~\citep{Reimers2019SentenceBERTSE}
to retrieve examples from the training set
can substantially enhance the performance of in-context learning over random selection.
Another approach involves training dense retrievers
based on the feedback signals from LLMs,
which has shown promising results in semantic parsing ~\citep{rubin2022learning},
cross-task prompt retrieval ~\citep{Cheng2023UPRISEUP},
and unified multi-task retrieval ~\citep{li2023unified}.
However,
existing methods either focus on a relatively small language model ~\citep{rubin2022learning},
or fail to exploit the fine-grained feedback information from LLMs in a principled manner ~\citep{li2023unified}.

In this paper,
we propose a novel framework, LLM-R (\textbf{LLM R}etriever),
which aims to retrieve high-quality in-context examples for large language models.
Given an initial set of retrieved candidates,
our framework ranks them based on the conditional LLM log probabilities of the ground-truth outputs.
Subsequently,
a cross-encoder based reward model is trained to capture the fine-grained ranking signals from LLMs.
Finally, a bi-encoder based dense retriever is trained using knowledge distillation.
The reward model plays a crucial role in providing more informative soft-labels that are suitable for distillation,
instead of using heuristically constructed one-hot labels.
This pipeline can be iterated multiple times by
retrieving a new set of candidates based on the latest dense retriever.

For evaluation purposes,
we assemble a diverse set of $30$ NLP tasks,
which span $9$ categories,
including question answering, natural language inference,
commonsense reasoning, and summarization, among others.
Experimental results obtained using LLaMA-7B ~\citep{Touvron2023LLaMAOA} demonstrate that
our model improves the in-context learning performance by an average of $7.8\%$
compared to random selection.
Similar improvements are also observed on held-out tasks and LLMs of varying sizes.
Further analysis reveals that the top-retrieved examples share
similar input patterns or the same labels as the testing example.
Our model is particularly effective for classification tasks with ample training examples.
In contrast,
tasks such as closed-book question answering and commonsense reasoning
rely more on the inherent capabilities of LLMs and
are less sensitive to the quality of in-context examples.

\section{Related Work}
\noindent
\textbf{In-Context Learning }
is an emergent property of large language models (LLMs) that
enables them to perform various tasks conditioned on a few input-output examples,
without any parameter updates or fine-tuning.
This property has been demonstrated in LLMs such as GPT-3 ~\citep{NEURIPS2020_1457c0d6}, GPT-Neo ~\citep{gpt-neo}, and LLaMA ~\citep{Touvron2023LLaMAOA},
and attracts considerable attention from the research community.
One area of research is focused on understanding the underlying mechanism and principles of in-context learning.
For instance, ~\citeauthor{xie2021explanation} view in-context learning as implicit Bayesian inference,
while ~\citeauthor{dai2022can} interpret it as meta optimization.

Another area of research is to explore different strategies for selecting and designing in-context examples for LLMs.
Recent studies ~\citep{liu2022makes,rubin2022learning,li2023unified,luo2023dr} have shown that
using BM25 algorithm or fine-tuning dense retrievers based on LLM feedback to retrieve from the training set
can improve the performance of in-context learning.
Our work also falls into this area by proposing a novel training method.
To model the interaction between in-context examples,
determinantal point process ~\citep{ye2023compositional} and sequential decision-making ~\citep{zhang2022active}
are introduced as preliminary explorations.
In contrast,
Structured Prompting ~\citep{hao2022structured} breaks the limitation of input context length
and scales the number of in-context examples to thousands.
\newline

\noindent
\textbf{Dense Retrieval }
is a widely used information retrieval approach that
utilizes dense vectors to perform semantic matching between queries and documents
in the latent space ~\citep{Reimers2019SentenceBERTSE,Wang2022TextEB}.
Compared to sparse retrieval methods such as BM25,
dense retrieval exploits the powerful modeling capacity of pre-trained language models (PLMs) ~\citep{kenton2019bert}
to learn relevance functions and
has the potential to overcome the vocabulary mismatch problem.
Various techniques such as hard negative mining ~\citep{Karpukhin2020DensePR},
knowledge distillation ~\citep{ren2021rocketqav2}, and continual pre-training ~\citep{Wang2022TextEB}
have been proposed to enhance the performance of dense retrieval.
\newline

\noindent
\textbf{Retrieval Augmented LLMs }
combine the generative power of LLMs with the ability to retrieve relevant information
from external sources ~\citep{ram2023context,lewis2020retrieval,shi2023replug}.
This paradigm has the potential to enhance the factual consistency of generated texts,
make LLMs aware of the up-to-date knowledge,
as well as provide a natural way for source attribution ~\citep{nakano2021webgpt}.
The retrieved information can be incorporated into LLMs through various mechanisms,
such as input concatenation ~\citep{shi2023replug},
intermediate attention fusion ~\citep{borgeaud2022improving},
and output interpolation ~\citep{khandelwal2019generalization}.
For in-context learning,
the goal of retrieval augmentation is to improve the performance of LLMs on downstream tasks
by retrieving informative examples ~\citep{li2023unified,luo2023dr}.

\begin{figure*}[ht]
\centering
\includegraphics[width=1.0\textwidth]{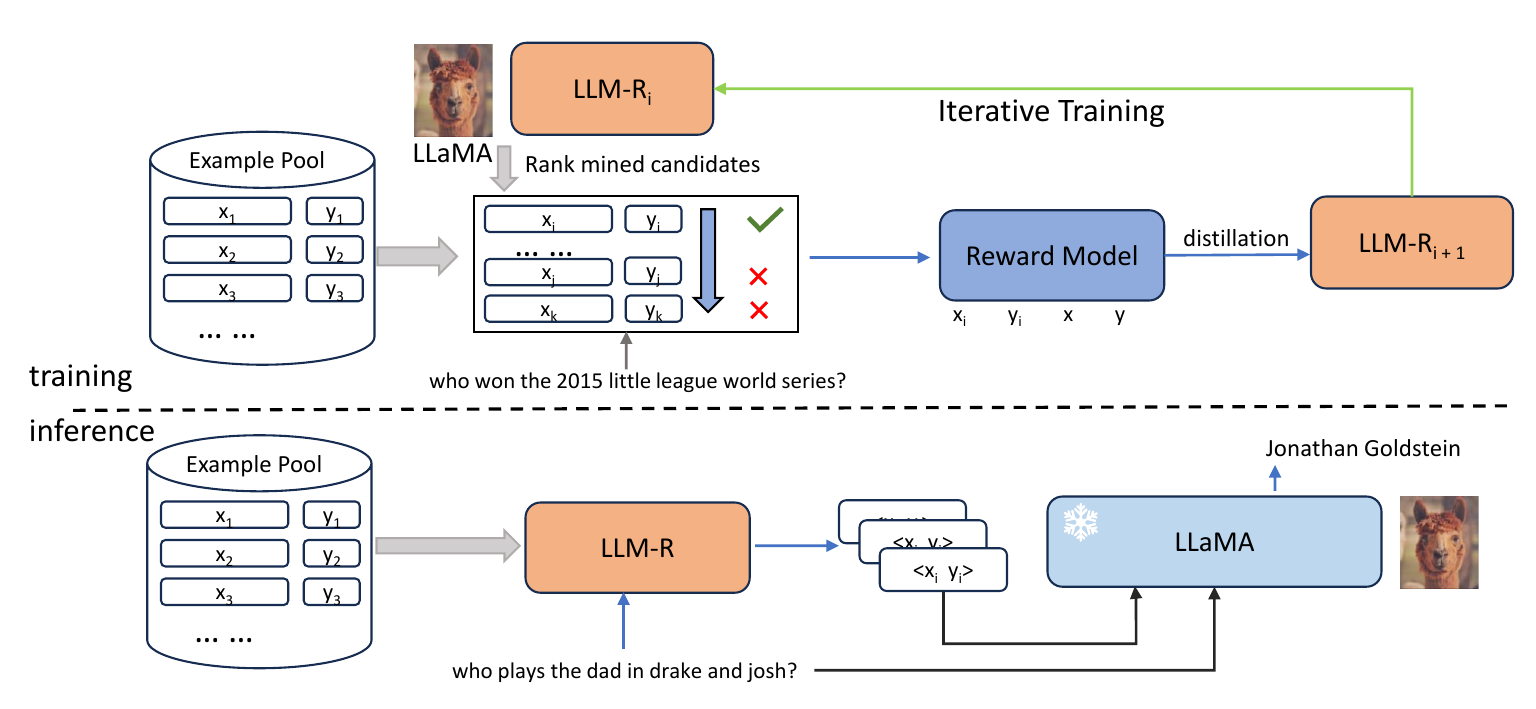}
\caption{The overall architecture of our proposed framework LLM-R.
The training process comprises three stages:
generating training data based on an initial retriever and LLM feedback,
reward modeling,
and training dense retrievers by distilling the knowledge from the reward model.
At inference time,
the trained dense retriever is employed to retrieve in-context examples from the pool $\mathbb{P}$
and the retrieved examples are fed to the LLM to generate the output.}
\label{fig:architecture}
\end{figure*}

\section{Preliminaries}
In this section,
we provide a brief introduction to the problem setting of in-context example retrieval.
Given a test example $x_\text{test}$ from a target task
and $k$ in-context examples $\{(x_i, y_i)\}_{i=1}^{k}$ from a pre-defined pool $\mathbb{P}$,
a frozen language model $M$ is employed to predict an output $y'_\text{test}$ through autoregressive decoding.
The primary objective of in-context example retrieval is to retrieve $k$ examples from $\mathbb{P}$
such that the predicted output $y'_\text{test}$ is as close as possible to the ground-truth output $y_\text{test}$
based on some task-specific metrics.
In this paper,
the example pool $\mathbb{P}$ is the union of the training set for all the tasks in our evaluation.

Straightforward solutions include utilizing the BM25 algorithm
or readily available text embedding models ~\citep{Wang2022TextEB,liu2022makes} to retrieve examples from $\mathbb{P}$
by treating $x_\text{test}$ as a query.
Despite their simplicity,
these methods have been shown to be more effective empirically
when compared to the random selection baseline.
In contrast,
our framework aims to learn a dense retriever customized for in-context example retrieval
by leveraging the feedback from LLMs.

\section{Methodology}

Our proposed framework is depicted in Figure ~\ref{fig:architecture}.
It includes four main components:
training data generation, reward modeling, dense retriever training, and inference,
which are described in detail in the following subsections.

\subsection{Training Data Generation}

\noindent
\textbf{Initial Candidates Retrieval }
Given an example $(x, y)$ from the training set,
where $x$ is the input and $y$ is the groundtruth output,
we retrieve the top-$n$ candidates $\{(x_i, y_i)\}_{i=1}^{n}$ from the example pool $\mathbb{P}$
using an initial retriever.
The pool $\mathbb{P}$ contains the training examples from a mixture of tasks.
Since $(x, y) \in \mathbb{P}$ holds during training,
we exclude itself from the retrieval results.

In this paper,
we employ the unsupervised BM25 algorithm as the initial retriever.
The query only consists of the input $x$,
while each retrieval candidate is the string concatenation of the input $x_i$ and the output $y_i$.
This setting aligns with the test-time scenario,
where the groundtruth output is unavailable.
With a reasonably effective initial retriever,
the top-$n$ candidates would likely contain some positive examples
and hard negative examples.
\newline

\noindent
\textbf{Ranking Candidates using LLMs }
To assess the quality of the retrieved candidates,
we utilize feedback signals from a frozen LLM.
Specifically,
we rank the candidates in descending order based on the log-likelihood of the groundtruth output $y$,
as given by the following equation:
\begin{equation}
\label{eq:llm_ranking}
    \log p(y | x, x_i, y_i), \forall i \in \{1, 2, \dots, n\}
\end{equation}
Here, $p(y | x, x_i, y_i)$ is the conditional probability of $y$
given the input $x$ and the $i$-th candidate $(x_i, y_i)$.
It is noteworthy that computing $p(y | x, x_i, y_i)$ requires only one forward pass,
and does not rely on any task-specific metrics,
despite the autoregressive nature of language models.
In practical applications,
this helps reduce the inference cost of LLMs.

\subsection{Reward Modeling}
In order to capture the preferences of LLMs over the retrieved candidates
and provide fine-grained supervision for dense retrievers,
we propose to train a cross-encoder based reward model.
For a training example $(x, y)$,
we first sample one positive example $(x^+, y^+)$ from the top-ranked candidates
and $N_\text{neg}$ hard negative examples $\{(x_i^-, y_i^-)\}_{i=1}^{N_\text{neg}}$ from the bottom-ranked candidates.
The reward model takes as input the concatenation of $(x, y, x^+, y^+)$
and produces a real-valued score $s(x, y, x^+, y^+)$,
similarly for the hard negatives.
It is trained to minimize the following cross-entropy loss:
\begin{equation}
\label{eq:reward_loss}
    \mathcal{L}_\text{reward} = -\log \frac{e^{s(x, y, x^+, y^+)}}{e^{s(x, y, x^+, y^+)} + \sum_{i=1}^{N_\text{neg}} e^{s(x, y, x_i^-, y_i^-)}}
\end{equation}
It is important to note that
the reward model is only used to provide supervision for the dense retriever
and has access to the groundtruth label $y$,
which is not available at test time.
This is a key difference from the re-ranker in the ad-hoc retrieval setting ~\citep{ren2021rocketqav2}.
Compared to the bi-encoder based dense retrievers,
the reward model enables full interaction between the inputs
and can therefore serve as a teacher model.
The log-likelihood scores from LLMs display high variance across different examples.
In contrast,
the reward model scores are more suitable for knowledge distillation.

\subsection{Training LLM Retrievers with Knowledge Distillation}
To facilitate efficient inference,
the dense retriever is based on the bi-encoder architecture.
Given a query $x$,
we compute its low-dimensional embedding $\mathbf{h}_x$
by performing average pooling over the last-layer hidden states.
Similarly,
we obtain the embedding $\mathbf{h}_{(x_i, y_i)}$ for the candidate $(x_i, y_i)$
by taking the concatenation of $x_i$ and $y_i$ as input.
The matching score $f(x, x_i, y_i)$ is computed as
the temperature-scaled cosine similarity
$\cos(\mathbf{h}_x, \mathbf{h}_{(x_i, y_i)}) / \tau$,
where $\tau$ is a temperature hyperparameter.
In this paper,
we use a shared encoder for both the query and the retrieval candidates.

The dense retriever is trained to distill the knowledge from the reward model.
We use the KL divergence loss $\mathcal{L}_\text{distill}\ = \ \text{KL}(p_\text{reward}\ ||\ p_\text{retriever})$
to measure the mismatch between the reward model distribution $p_\text{reward}$
and the retriever distribution $p_\text{retriever}$.
$\mathcal{L}_\text{distill}$ is only computed over the hard negatives for efficiency reasons.
To incorporate the in-batch negatives,
we also include an InfoNCE-based contrastive loss $\mathcal{L}_\text{cont}$ ~\citep{Chen2020ASF}
by treating the candidate with the highest reward as the positive example.
The final loss function $\mathcal{L}_\text{retriever}$ is a weighted sum of
the contrastive loss and the knowledge distillation loss:
\begin{equation}
\label{eq:retriever_loss}
    \mathcal{L}_\text{retriever}\ =\ \alpha \mathcal{L}_\text{cont}\ +\ \mathcal{L}_\text{distill}
\end{equation}
Here, $\alpha$ is a constant that controls the relative importance of the two losses.
\newline

\noindent
\textbf{Iterative Training }
As illustrated in Figure ~\ref{fig:architecture},
the retriever trained in iteration $i$
can be employed to retrieve candidates for the subsequent iteration $i+1$.
In the first iteration,
the candidates are retrieved using BM25.
Such an iterative training approach ~\citep{xiong2020approximate,li2023unified} allows
improving retriever quality by mining better positive and hard negative examples.

\subsection{Evaluation of LLM Retrievers}
Given a test example $x_\text{test}$,
we compute its embedding $\mathbf{h}_\text{test}$ using the trained retriever
and retrieve the top $k$ candidates from the pool $\mathbb{P}$ as the $k$-shot in-context examples.
The input to the LLM is the concatenation of the $k$-shot examples and $x_\text{test}$.
The overall procedure is illustrated in Figure ~\ref{fig:architecture}.

Depending on the task type of $x_\text{test}$,
different decoding strategies are employed to generate the final prediction.
For classification tasks,
we use greedy search with constrained decoding to make sure the prediction is a valid class label.
For multiple choice tasks,
all the choices are ranked based on the average token-level log-likelihood score,
and the one with the highest score is selected as the model's prediction.
Generation tasks use greedy search without any constraints.
For quantitative evaluation,
the prediction is compared with the groundtruth $y_\text{test}$ using task-specific metrics.

\section{Experiments}

\begin{figure*}[ht]
\centering
\includegraphics[width=1.0\textwidth]{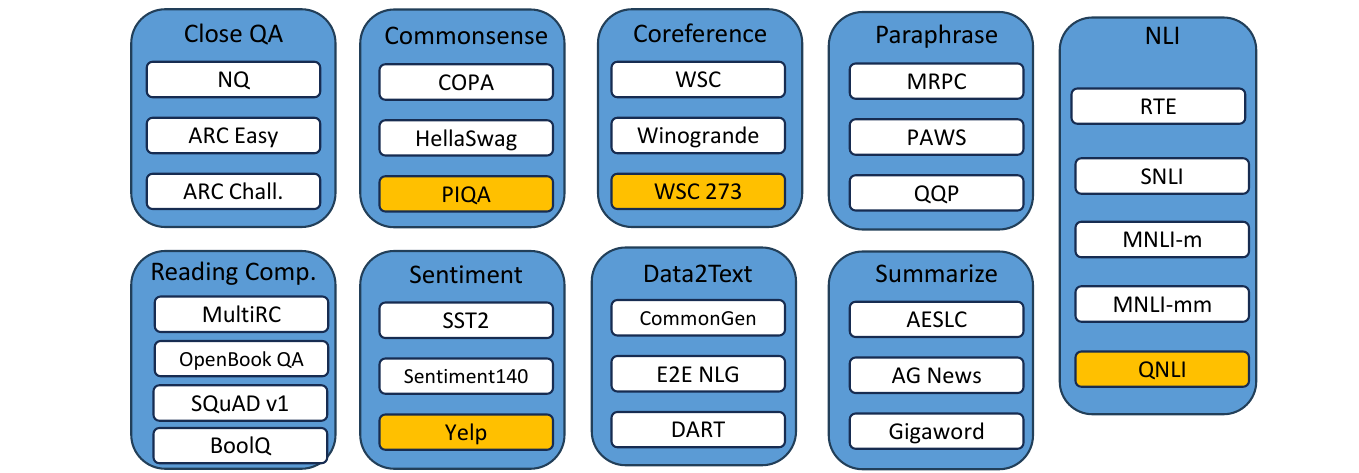}
\caption{The collection of datasets used in our experiments.
The yellow-colored datasets are held out and excluded from training.
For further information,
please refer to Table ~\ref{tab:datasets_stat} in the Appendix.}
\label{fig:datasets}
\end{figure*}

\begin{table*}[ht]
\centering
\scalebox{0.95}{\begin{tabular}{lcccccccccc}
\hline
\multicolumn{1}{l}{\multirow{2}{*}{\# of datasets $\rightarrow$}}  & CQA & Comm. & Coref. & NLI & Para. & RC & Sent. & D2T & Summ. & Avg \\
\multicolumn{1}{l}{}  & 3  & 3  & 3  & 5  & 3  & 4  & 3  & 3 & 3 & 30  \\ \hline
Zero-shot &   29.0 & 71.5 & 66.8 & 44.0 & 60.0 & 41.3 & 50.5 & 25.6 & 17.5 & 44.9 \\
Random  &  40.4 &	77.6 &	67.2 & 50.9 & 56.6 & 58.1 & 88.8 & 47.0 & 38.9  &  57.9   \\
K-means & 41.6 & 79.5 & 66.0 & 50.8 & 52.6 & 53.6 & 90.9 & 42.5 & 40.5 & 57.0  \\
BM25  &    45.9 & 78.1 & 62.9 & 54.7 & 66.1 & 59.9 & 89.6 & 49.3 & 50.0 &  61.3  \\
E5$_{\text{base}}$  &  \textbf{49.0} & 79.8 & 64.6 & 53.6 & 58.0 & 60.2 & \textbf{94.4} & 48.0 & 50.0 & 61.4   \\
SBERT  & 48.5 & 79.3 & 64.2 & 57.5 & 64.1 & \textbf{60.6} & 91.9 & 47.4 & 49.3 & 62.1  \\
EPR$^\dagger$ & 48.4 & 79.3 & 64.4 & 64.3 & 65.1 & 59.8 & 91.7 & 49.7 & 50.0 & 63.5 \\  \hline
LLM-R (1 iter) &  48.8 & 80.1 & 67.6 & 71.9 & 66.5 & 60.0 & 93.5 & \textbf{50.1} & 50.8 & 65.7    \\
LLM-R (2 iter) & 48.7 & \textbf{80.4} & 70.4 & 72.5 & 71.5 & 59.0 & 93.6 & 49.9 & \textbf{51.1} & \textbf{66.5}  \\
LLM-R (3 iter)  & 48.9 & 80.0 & \textbf{70.8} & \textbf{72.6} & \textbf{72.8} & 58.0 & 92.9 & 49.8 & 50.8 & 66.4  \\ \hline
Std dev.  & $\pm$0.2 & $\pm$0.8 & $\pm$0.7 & $\pm$0.1 & $\pm$1.1 & $\pm$0.0 & $\pm$0.4 & $\pm$0.0 & $\pm$0.1 & $\pm$0.2   \\ \hline
\end{tabular}}
\caption{Our main results.
We report the average metrics for Close QA (CQA), Commonsense Reasoning (Comm.),
Coreference (Coref.), NLI, Paraphrase (Para.), Reading Comprehension (RC),
Sentiment (Sent.), Data-to-text (D2T), Summarize (Summ.).
The standard deviation is computed over $3$ runs with the ``Random'' baseline.
Dense retriever baselines include E5 ~\citep{Wang2022TextEB}, SBERT ~\citep{Reimers2019SentenceBERTSE},
and EPR ~\citep{rubin2022learning}.
$^\dagger$: Our re-implementation for fair comparison.}
\label{tab:main_results}
\end{table*}

\subsection{Evaluation Setup} \label{subsec:exp_setup}
We utilize a total of $30$ publicly available datasets ~\footnote{We use ``datasets'' and ``tasks'' interchangeably.}
from $9$ distinct categories for training and evaluation,
as shown in Figure ~\ref{fig:datasets}.
This collection is based on FLAN ~\citep{wei2021finetuned} and UPRISE ~\citep{Cheng2023UPRISEUP}.
Different from our work,
FLAN is focused on fine-tuning language models to follow instructions,
while UPRISE is designed for cross-task retrieval.
To test the generalization ability of the models to unseen tasks,
we held out four datasets,
namely QNLI, PIQA, WSC273, and Yelp,
from the training process.
The retrieval pool is created by taking the union of all the training examples,
which results in a total of approximately $6.3$M examples.
For each dataset,
we sample a maximum of $30$k examples for training
and $10$k examples for evaluation to reduce the cost of LLM inference.
For evaluation,
we report the average metrics in each task category.
Please check Table ~\ref{tab:datasets_stat} for the specific metrics used for each dataset.

In the main experiments,
we use LLaMA-7B ~\citep{Touvron2023LLaMAOA} as the default LLM for candidate ranking and task evaluation
unless otherwise specified.
The reward model is initialized with ELECTRA$_\text{base}$ ~\citep{clark2019electra}
and the retriever is initialized with E5$_\text{base}$ ~\citep{Wang2022TextEB}.
The baselines include zero-shot prompting, k-means clustering, random selection, BM25 ~\citep{Lin2021PyseriniAP},
and two off-the-shelf dense retrievers,
namely SBERT (all-mpnet-base-v2) ~\citep{Reimers2019SentenceBERTSE} and E5$_\text{base}$.
Except for zero-shot evaluation,
we retrieve $8$ in-context examples for each test input.
More implementation details and training hyperparameters are in Appendix ~\ref{sec:app_implementation_details}.

\begin{table*}[ht]
\centering
\scalebox{0.89}{\begin{tabular}{lcccccccccc}
\hline
                 & CQA & Comm. & Coref. & NLI & Para. & RC & Sent. & D2T & Summ. & Avg \\ \hline
LLM-R (1 iter)  &  \textbf{48.8} & \textbf{80.1} & 67.6 & \textbf{71.9} & 66.5 & 60.0 & \textbf{93.5} & \textbf{50.1} & \textbf{50.8} & \textbf{65.7}   \\ \hline
\multicolumn{11}{l}{\emph{model variants}}                  \\
\ \ w/o reward model & \textbf{48.8} & 79.1 & 64.3 & 68.9 & 70.2 & 60.5 & 91.7 & 49.4 & 50.5 & 64.9$^{\downarrow0.8}$  \\
\ \ reward model w/o label $y^+$ & 48.5 & 79.7 & 67.5 & 64.1 & 62.7 & \textbf{60.8} & 92.3 & 49.6 & 49.8 & 63.8$^{\downarrow1.9}$  \\
\ \ LLM score as reward & 48.0 & 79.4 & 67.0 & 67.0 & \textbf{74.0} & 60.5 & 91.5 & 49.6 & 50.3 & 65.2$^{\downarrow0.5}$ \\ \hline
\multicolumn{11}{l}{\emph{retriever initialization}}                  \\
\ \ initialize w/ BERT$_{\text{base}}$ & 48.7 & 79.6 & \textbf{69.4} & 70.9 & 63.0 & 60.7 & 92.0 & 50.0 & 50.2 & 65.2$^{\downarrow0.5}$ \\ \hline
\end{tabular}}
\caption{Different training variants of LLM-R.
``w/o reward model'' is trained solely with contrastive loss on LLM ranked candidates.
``LLM score as reward'' uses the log-likelihood score from LLMs as the distillation target.
Neither of these two variants utilizes the reward model.
``reward model w/o label $y^+$'' denotes that the reward model is trained without access to the groundtruth label $y^+$.}
\label{tab:pipeline_ablation}
\end{table*}

\begin{table*}[ht]
\centering
\scalebox{0.95}{\begin{tabular}{lccccccc}
\hline
 & Zero-shot & Random & K-means & BM25 & E5$_{\text{base}}$ & SBERT & LLM-R \\ \hline
QNLI & 49.2 & 56.4 & 53.4 & 62.2 & 61.5 & 61.9 & \textbf{69.6}$^{\uparrow7.7}$ \\
PIQA & 77.0 &  79.1 & 79.4 & 81.3 & 81.3 & 80.7 & \textbf{81.6}$^{\uparrow0.3}$ \\
WSC273 & 74.0 & 74.4 & 74.7 & 64.5 & 65.2 & 62.6 & \textbf{79.5}$^{\uparrow4.8}$ \\
Yelp & 47.9 & 92.0 & 93.5 & 93.5 & \textbf{97.3} & 95.9 & 95.9$^{\downarrow1.4}$ \\ \hline
Average & 62.0 & 75.5 & 75.3 & 75.4 & 76.3 & 75.3 & \textbf{81.7}$^{\uparrow5.4}$ \\ \hline
\end{tabular}}
\caption{Generalization to four held-out tasks.}
\label{tab:generalize_unseen_tasks}
\end{table*}

\subsection{Main Results}
Table ~\ref{tab:main_results} presents the main results of our experiments.
We observe that the simple BM25 algorithm serves as a strong baseline,
exhibiting consistent improvements over the random selection strategy.
This conclusion aligns with the findings of ~\citeauthor{luo2023dr}.
Such effectiveness of BM25 can help warm up the first training iteration
by providing a set of high-quality candidates.
We also tried to use E5$_\text{base}$ as the initial retriever,
but the benefits compared to BM25 are marginal.
Therefore,
we stick to BM25 for its simplicity.

After the first iteration,
our proposed model LLM-R outperforms all the baselines ($63.5 \rightarrow 65.7$)
by training on the BM25 retrieved candidates.
The second iteration includes the mined positive and hard negative examples from ``LLM-R (1 iter)'',
raising the average score to $66.5$ (+$0.8$).
Further iterations do not yield substantial improvements,
indicating that the model has converged.

\section{Analysis}
In this section,
we examine the performance of LLM-R across various tasks, LLMs, and model variants.
Unless explicitly specified,
``LLM-R'' refers to the model with $2$ training iterations.

\subsection{Training Pipeline of LLM-R}

We investigate several LLM-R variants LLM-R in Table ~\ref{tab:pipeline_ablation} to understand the contribution of each component.
The ``w/o reward model'' variant removes the knowledge distillation loss
and sees $0.8$ points drop in average score.
This indicates that the reward model is crucial for the performance of LLM-R.
Inspired by REPLUG ~\citep{shi2023replug},
we experiment with a variant that uses the log-likelihood from LLMs as the reward for distillation.
Although it outperforms the ``w/o reward model'' variant,
it still lags behind our method by $0.5$ points.
The average token-level log-likelihood from LLMs is not a probability distribution by nature.
We empirically observe that feedback scores for some training examples are concentrated in a very narrow range,
while other scores are more dispersed.
This makes it suboptimal to serve as target distribution within KL-divergence framework.
Changing the retriever initialization from E5 ~\citep{Wang2022TextEB} to BERT ~\citep{kenton2019bert}
results in a performance drop,
but not as significant as in the ad-hoc retrieval setting.

\subsection{Generalization Ability of LLM-R}
We evaluate the generalization ability of LLM-R from two dimensions.
In the first scenario,
we test whether the trained retriever can retrieve good in-context examples
for tasks that are not seen during training.
In the second scenario,
we test whether a model trained with one LLM can generalize to other LLMs
that vary in size and quality.

In Table ~\ref{tab:generalize_unseen_tasks},
we report the performance of LLM-R on four held-out tasks.
The results demonstrate that
LLM-R surpasses the second-best model E5$_\text{base}$ by an average of $5.4$ points,
indicating its ability to generalize to previously unseen tasks.
Under the current evaluation protocol,
there are training datasets that share the same task category
as the held-out ones (e.g., QNLI and SNLI are both for natural language inference).
A more challenging setting is to test on non-overlapping task categories,
which we leave for future work.

\begin{table*}[ht]
\centering
\scalebox{0.88}{\begin{tabular}{lcccccccccl}
\hline
  & CQA & Comm. & Coref. & NLI & Para. & RC & Sent. & D2T & Summ. & Avg \\ \hline
\multicolumn{11}{l}{\emph{gpt-neo-2.7b}} \\
\ \ BM25 & 41.1 & 67.0 & 53.2 & 47.6 & 64.5 & 51.2 & 78.3 & 45.4 & 47.3 & 54.4 \\
\ \ LLM-R & 42.2 & 68.0 & 59.7 & 71.5 & 73.0 & 51.6 & 91.6 & 46.9 & 48.8 & \textbf{61.8}$^{\uparrow7.4}$ \\ \hline
\multicolumn{11}{l}{\emph{llama-13b}}    \\
\ \ BM25 & 49.6 & 80.1 & 61.1 & 67.0 & 69.9 & 60.5 & 92.5 & 49.9 & 50.9 & 64.6 \\
\ \ LLM-R & 52.0 & 83.7 & 71.2 & 76.8 & 73.3 & 62.2 & 94.2 & 50.7 & 52.0 & \textbf{68.8}$^{\uparrow4.2}$ \\ \hline
\multicolumn{11}{l}{\emph{gpt-35-turbo$^\dagger$}} \\
\ \ BM25  & 75.3 & 85.2 & 65.0 & 78.1 & 78.0 & 84.4 & 95.7 & 51.9 & 52.8 & 74.7 \\
\ \ LLM-R  & 79.3 & 86.7 & 63.8 & 79.6 & 76.0 & 84.0 & 95.4 & 52.2 & 53.0 & \textbf{75.1}$^{\uparrow0.4}$ \\ \hline
\end{tabular}}
\caption{Generalization to LLMs that are not used for training.
$\dagger$: Since the official API of \emph{gpt-35-turbo} does not return the log-probabilities,
we use different input-output templates to formulate all tasks as text generation.
Consequently,
the scores of \emph{gpt-35-turbo} cannot be directly compared with those of other LLMs.
More details are in Appendix ~\ref{sec:app_gpt35turbo}.}
\label{tab:generalize_other_llms}
\end{table*}

\begin{figure*}[ht]
\begin{center}
 \includegraphics[width=0.95\linewidth]{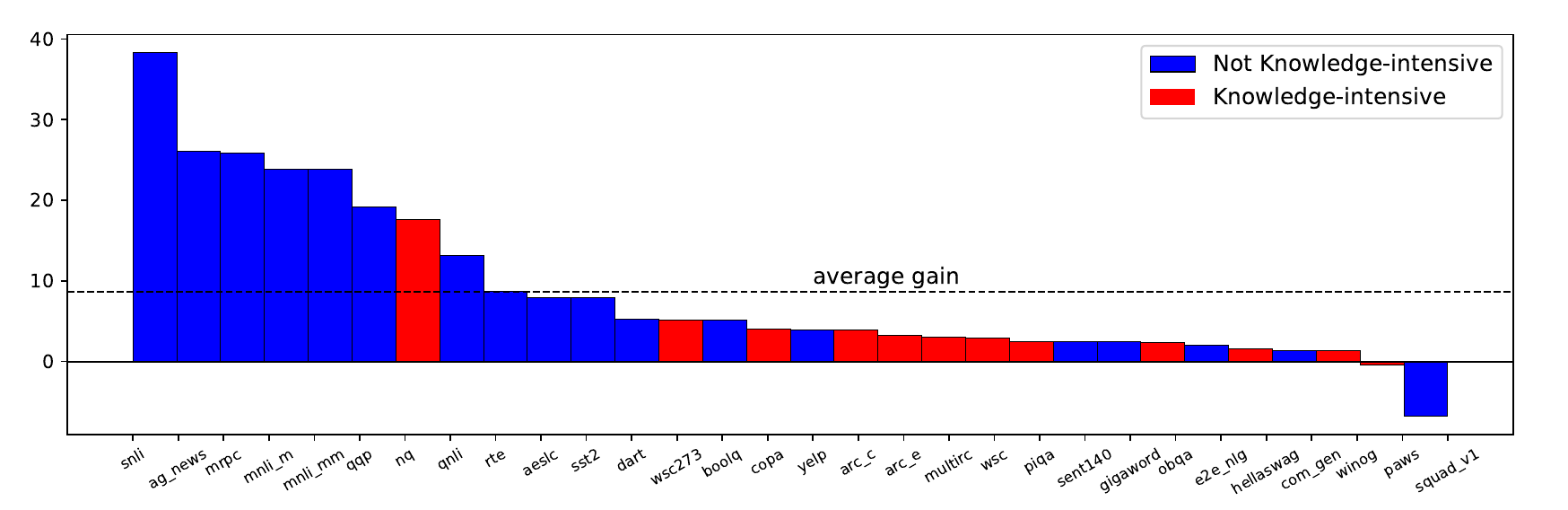}
 \caption{Performance gains of LLM-R over the random selection baseline.
    The selected \emph{knowledge-intensive} tasks are NQ, ARC (easy and challenge), PIQA,
    HellaSwag, COPA, Paws, OpenBook QA, WSC273, WSC, Winogrande, and MultiRC.}
 \label{fig:performance_gains}
\end{center}
\end{figure*}

The LLM-R model is trained with LLaMA-7B.
To evaluate its generalization ability across different LLMs,
we test on three other models,
namely GPT-Neo-2.7B ~\citep{gpt-neo}, LLaMA-13B, and GPT-35-Turbo.
Results in Table ~\ref{tab:generalize_other_llms} show that
LLM-R consistently outperforms the BM25 baseline for LLMs
with parameter ranges from $2.7$B to tens of billions.
Notably,
the gains are particularly significant for small-size language models,
possibly because they are less powerful and
thus require higher-quality examples to perform in-context learning.

\subsection{When does LLM-R Work and When Does it Not?}

Reporting a single aggregate score for all tasks facilitates comparison across different model variants.
However,
this approach hides the fact that LLM-R performs better on certain tasks than others,
and may even lead to performance degradation in some cases.
In Figure ~\ref{fig:performance_gains},
we partition the tasks into two groups.
A task is considered to be \emph{knowledge-intensive} if solving this task requires
commonsense, complex reasoning, or memorized factual knowledge.

For tasks in the knowledge-intensive set,
the absolute improvements are substantially smaller than the average,
with NQ being the only exception.
This is not surprising,
as these tasks rely more heavily
on the underlying foundation model's capability to perform reasoning and knowledge memorization.
For the NQ dataset,
we empirically find that there is some overlap between the training and test sets,
where test questions are paraphrases of some training questions.
Despite this,
we decide to keep the NQ dataset in our evaluation,
as it is a widely used benchmark and
the remaining non-overlapping questions are still valuable.

\begin{table*}[ht]
\centering
\scalebox{0.9}{\begin{tabular}{ll}
\hline
\small Task name  & \begin{tabular}[c]{@{}p{0.95\linewidth}@{}} \small Sentiment140 \end{tabular} \\
\small Test Input  & \begin{tabular}[c]{@{}p{0.95\linewidth}@{}} \small Math review. Im going to fail the exam. What is the sentiment of this tweet? \end{tabular} \\
\small Test Answer  & \begin{tabular}[c]{@{}p{0.95\linewidth}@{}} \small \textbf{Negative} \end{tabular} \\
\small LLM-R & \begin{tabular}[c]{@{}p{0.95\linewidth}@{}} \small revising for maths exam on tuesday which im gonna fail badly  What is the sentiment of this tweet? \textbf{Negative} \end{tabular} \\ \hline
\small Task name  & \begin{tabular}[c]{@{}p{0.95\linewidth}@{}} \small MNLI-m \end{tabular} \\
\small Test Input  & \begin{tabular}[c]{@{}p{0.95\linewidth}@{}} \small Premise: "Part 2), Confidentiality of Alcohol and Drug Abuse Patient Records." Hypothesis: "Drug and alcohol patient records should be confidential" Does the premise entail the hypothesis? Yes, No, or Maybe? \end{tabular} \\
\small Test Answer  & \begin{tabular}[c]{@{}p{0.95\linewidth}@{}} \small \textbf{Yes} \end{tabular} \\
\small LLM-R & \begin{tabular}[c]{@{}p{0.95\linewidth}@{}} \small Premise: "Eligible Clients unable to attain needed legal assistance" Hypothesis: "Clients that should have received legal assistance but didn't" Does the premise entail the hypothesis? Yes, No, or Maybe? \textbf{Yes} \end{tabular} \\ \hline
\small Task name  & \begin{tabular}[c]{@{}p{0.95\linewidth}@{}} \small PIQA \end{tabular} \\
\small Test Input  & \begin{tabular}[c]{@{}p{0.95\linewidth}@{}} \small Here is a goal: "How can I keep a bathroom mirror from fogging up?" How would you accomplish this goal? \end{tabular} \\
\small Test Answer  & \begin{tabular}[c]{@{}p{0.95\linewidth}@{}} \small \textbf{Wipe down with shaving cream.} \end{tabular} \\
\small LLM-R & \begin{tabular}[c]{@{}p{0.95\linewidth}@{}} \small Here is a goal: "how do you 'clean up' an eyebrow you've filled in?" How would you accomplish this goal? \textbf{use concealer to cover up any mistakes made.} \end{tabular} \\ \hline
\end{tabular}}
\caption{Retrieved examples by LLM-R.
The bold texts are the groundtruth answers for the test inputs and retrieved candidates.
More examples are available in Table ~\ref{tab:app_retrieved_examples}.}
\label{tab:cases}
\end{table*}

Another noticeable case is the SQuAD v1 dataset ~\citep{rajpurkar2016squad},
where LLM-R performs worse than the random selection baseline.
Upon manual inspection,
we find that many questions in SQuAD share the same passage as the context.
This frequently results in LLM-R retrieving examples with limited diversity,
which may account for the observed decline in performance.

In Table ~\ref{tab:cases},
for the Sentiment140 and MNLI datasets,
our model helps by retrieving examples that share similar input patterns with the test example.
In contrast,
the PIQA dataset requires commonsense knowledge and may not benefit much from the retrieved examples.

\subsection{Using Different LLMs for Data Generation and Task Evaluation}

\begin{table}[ht]
\centering
\scalebox{0.85}{\begin{tabular}{lccc}
\hline
\begin{tabular}[c]{@{}l@{}}Rank LLM\ $\rightarrow$ \\ Eval LLM\ $\downarrow$ \end{tabular} & GPT-Neo-2.7B & LLaMA-7B & Both \\ \hline
GPT-Neo-2.7B    &   \textbf{61.7}   &  61.3  &  61.6     \\
LLaMA-7B    &  66.0   & 65.7  &  \textbf{66.3}  \\ \hline
\end{tabular}}
\caption{On the impacts of using different LLMs for candidate ranking and task evaluation.
The ``Both'' setting merges the training data from two LLMs.}
\label{tab:feedback_evaluation}
\end{table}

One crucial aspect of our framework is the selection of the LLM for training data generation and task evaluation.
During the training phase,
the LLM plays a pivotal role in ranking the retrieved candidates and
providing supervision signals for the reward model.
In the task evaluation phase,
the LLM is used to generate the final predictions.

We experiment with GPT-Neo-2.7B and LLaMA-7B.
Table ~\ref{tab:feedback_evaluation} shows the results under
different combinations of LLMs for training and evaluation.
We observe that the quality of the evaluation LLM is the primary determinant for the final performance,
while the choice of ranking LLM has a relatively minor impact.
Although merging the training data from two LLMs yields the best overall performance,
we do not employ this technique in our main experiments for the sake of simplicity.

\subsection{Scaling the Number of In-Context Examples and Retriever Size}

\begin{figure}[ht]
\begin{center}
 \includegraphics[width=1.0\linewidth]{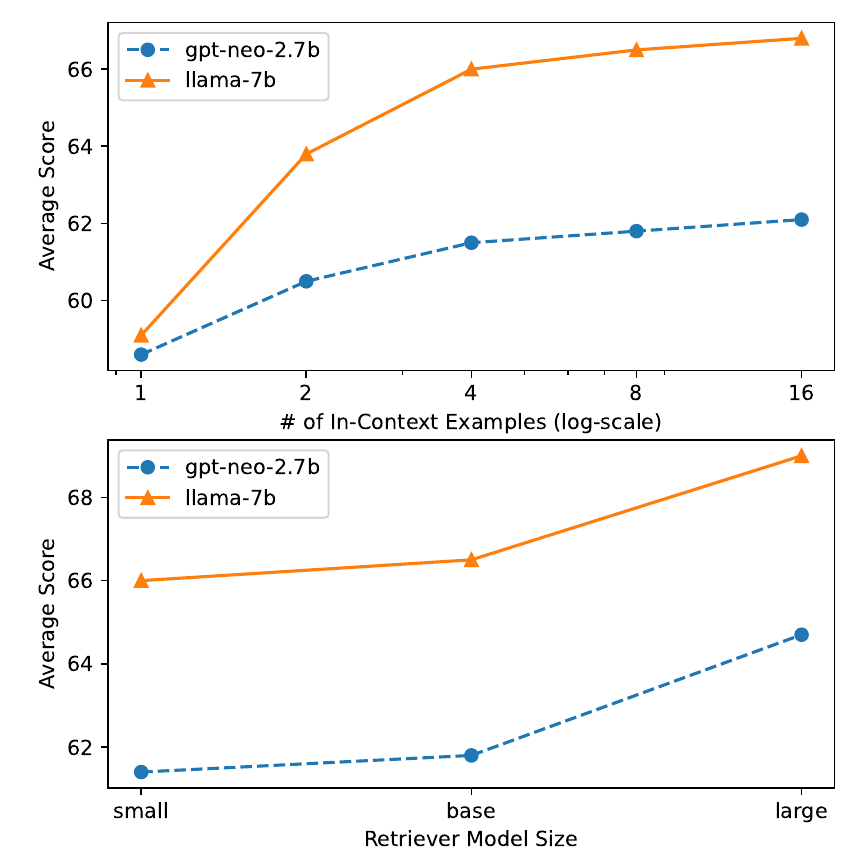}
 \caption{The scaling effect with respect to the number of in-context examples and retriever size.
 Our main experiments use $8$ in-context examples and base-size retriever.
 We vary the retriever model size by initializing with
 the released E5-\{small, base, large\} checkpoints from ~\citeauthor{Wang2022TextEB}.}
 \label{fig:scaling_icl}
\end{center}
\end{figure}

In Figure ~\ref{fig:scaling_icl},
we investigate the scaling effect of LLM-R from two aspects:
the number of in-context examples and the retriever model size.
The overall performance improves as we increase the number of retrieved examples,
but the gains diminish after $4$ examples.
Including more examples usually leads to longer prompts and higher inference cost.

With regard to the retriever size,
we observe that the small-size model produces comparable results with the base-size one,
whereas the large-size retriever exhibits a more substantial performance boost.
The trends are consistent for the two examined language models.
Practitioners can select the appropriate configurations based on the trade-off between performance and computational cost.

\section{Conclusion}
In this paper,
we introduce an iterative training framework named \emph{LLM-R}
to retrieve high-quality in-context examples for large language models.
This framework generates training data by utilizing a frozen LLM to rank the top retrieved candidates,
and then learns a cross-encoder based reward model to capture the ranking preference.
Bi-encoder based dense retrievers are trained to distill the knowledge from the reward model.
We conduct a comprehensive evaluation of LLM-R on a diverse set of tasks
and demonstrate that it consistently outperforms various strong baselines.
Our model also generalizes well to held-out tasks and LLMs of varying sizes.

\section*{Limitations}
In our framework,
we treat each candidate example independently and
retrieve the top-$k$ results for each test example.
This may be suboptimal as the in-context examples can influence each other.
Incorporating the techniques from the field of combinatorial optimization and sequential decision making
can be a promising direction to explore.

Another limitation of our study is related to the automatic evaluation protocol.
To compare the performance of different methods,
we report the arithmetic mean of the metrics over all tasks.
However,
this may put generation tasks at a disadvantage
since metrics like ROUGE and BLEU typically
have a narrower range of variation compared to classification accuracy.
Moreover,
the simple arithmetic mean does not account for the quality of each dataset.

\section*{Acknowledgements}
We would like to thank anonymous reviewers for their valuable comments,
and EACL 2024 and ACL Rolling Review organizers for their efforts.

\bibliography{custom}

\appendix

\section{Implementation Details} \label{sec:app_implementation_details}

\begin{table}[ht]
\centering
\scalebox{0.9}{\begin{tabular}{lcc}
\hline
 & Retriever & Reward Model \\ \hline
initialization &  E5$_\text{base}$    &  ELECTRA$_\text{base}$   \\
learning rate & $3\times10^{-5}$   &   $10^{-5}$  \\
\# of GPUs &  8  &    8          \\
batch size &   256   &  128   \\
train steps & 6k    &   3k     \\
$\tau$ &   0.01   &   n.a.    \\
$\alpha$ &  0.2     &   n.a.        \\
positive examples &  top 3  &  bottom 16  \\
negative examples &  top 3     &   bottom 16     \\
\# of negatives &  3     & 7   \\
ranking depth & 100 & 100 \\
input length &   256   &  384  \\ \hline
\end{tabular}}
\caption{Hyperparameters for training the bi-encoder retriever and reward model.
We use the same hyperparameters for every iteration.}
\label{tab:app_hyperparams}
\end{table}

\begin{table*}[ht]
\centering
\scalebox{0.85}{\begin{tabular}{lccccc}
\hline
Dataset name & Category & \# train & \# test  & Metric & Held-out? \\ \hline
AESLC ~\citep{zhang2019email}  &  Summarize  &  13,181  &  1,750 & ROUGE-L &  N  \\
AGNews ~\citep{zhang2015character}  &  Summarize  &  120,000  &  7,600 & Accuracy  &  N  \\
ARC Challenge ~\citep{bhakthavatsalam2021think}  &  Close QA  &  1,117  &  1,165 & Accuracy  &  N  \\
ARC Easy ~\citep{bhakthavatsalam2021think}  &  Close QA  &  2,241  &  2,365 & Accuracy  &  N  \\
BoolQ ~\citep{clark2019boolq}  &  Reading Comp.  &  9,427  &  3,270 & Accuracy  &  N  \\
CommonGen ~\citep{lin2020commongen}  &  Data-to-text  &  67,389  &  4,018 & ROUGE-L  &  N  \\
COPA ~\citep{roemmele2011choice}  &  Commonsense  &  400  &  100 & Accuracy  &  N  \\
DART ~\citep{nan2021dart}  &  Data-to-text  &  62,659  &  2,768 & ROUGE-L  &  N  \\
E2E NLG ~\citep{duvsek2019semantic}  &  Data-to-text  &  33,525  &  1,847 & ROUGE-L  &  N  \\
Gigaword ~\citep{napoles2012annotated}  &  Summarize  &  2,044,465  &  730 & ROUGE-L  &  N  \\
HellaSwag ~\citep{zellers2019hellaswag}  &  Commonsense  &  39,905  &  10,042 & Accuracy  &  N  \\
MNLI (m) ~\citep{williams2018broad}  &  NLI  &  392,702  &  9,815 & Accuracy  &  N  \\
MNLI (mm) ~\citep{williams2018broad}  &  NLI  &  392,702  &  9,832 & Accuracy  &  N  \\
MRPC ~\citep{dolan2005automatically}  &  Paraphrase  &  3,668  &  408 & Accuracy  &  N  \\
MultiRC ~\citep{khashabi2018looking}  &  Reading Comp.  &  27,243  &  4,848 & F1  &  N  \\
NQ ~\citep{Kwiatkowski2019NaturalQA}  &  Close QA  &  87,925  &  3,610 & Exact Match  &  N  \\
OpenBook QA ~\citep{mihaylov2018can}  &  Reading Comp.  &  4,957  &  500 & Accuracy  &  N  \\
PAWS ~\citep{zhang2019paws}  &  Paraphrase  &  49,401  &  8,000 & Accuracy  &  N  \\
PIQA ~\citep{bisk2020piqa}  &  Commonsense  &  16,113  &  1,838 & Accuracy  &  Y  \\
QNLI ~\citep{rajpurkar2018know}  &  NLI  &  104,743  &  5,463 & Accuracy  &  Y  \\
QQP ~\citep{wang2018glue}  &  Paraphrase  &  363,846  &  40,430 & Accuracy  &  N  \\
RTE ~\citep{bentivogli2009fifth}  &  NLI  &  2,490  &  277 & Accuracy  &  N  \\
Sentiment140 ~\citep{go2009twitter}  &  Sentiment  &  1,600,000  &  359 & Accuracy  &  N  \\
SNLI ~\citep{Bowman2015ALA}  &  NLI  &  549,367  &  9,824 & Accuracy  &  N  \\
SQuAD v1 ~\citep{rajpurkar2016squad}  &  Reading Comp.  &  87,599  &  10,570 & Exact Match  &  N  \\
SST2 ~\citep{Socher2013RecursiveDM}  &  Sentiment  &  67,349  &  872 & Accuracy  &  N  \\
Winogrande ~\citep{sakaguchi2021winogrande}  &  Coreference  &  40,398  &  1,267 & Accuracy  &  N  \\
WSC ~\citep{levesque2012winograd}  &  Coreference  &  554  &  104 & Accuracy  &  N  \\
WSC273 ~\citep{levesque2012winograd}  &  Coreference  &  0  &  273 & Accuracy  &  Y  \\
Yelp ~\citep{zhang2015character}  &  Sentiment  &  490,456  &  33,285 & Accuracy  &  Y  \\ \hline
Total  & n.a. &  6.3M    &  177k & n.a. & n.a.  \\
Total (sampled) & n.a.  &  591k     &  123k & n.a. & n.a.  \\ \hline
\end{tabular}}
\caption{Statistics for the datasets used in this paper.}
\label{tab:datasets_stat}
\end{table*}

\begin{table*}[ht]
\centering
\scalebox{0.85}{\begin{tabular}{ll}
\hline
Input  & \begin{tabular}[c]{@{}p{1.0\linewidth}@{}}What happens next in this paragraph? How to survive remedial classes Look at the course as an opportunity. Many students are discouraged when they are assigned to a remedial class. Some assume this placement means they aren't ready for college. OPTIONS: \\ A) However, people who are not unable to do what they're given on campus, or those who are cut out from college academies, are likely to have some little snitches. You want to be prepared for a negative outcome if possible. \\ B) In this case, you should consider what you will do if your subject consists of a certain term or number of subject areas. You could set up a study study program yourself or tutor a student who is struggling to thoroughly comprehend where they sat for homework. \\ C) If you take the course, you might find you feel highly motivated after passing the test. Try to develop a positive attitude towards the course so that you are not discouraged when you take your homework at the end of the day. \\ D) However, being assigned a remedial class doesn't mean that you are behind, just that you have an opportunity to receive better instruction and improve your skills in a subject that you have struggled with in the past. There is nothing unusual about being asked to attend a remedial course: two thirds of community college students take at least one remedial course. \end{tabular} \\ \hline
Output & D  \\ \hline
\end{tabular}}
\caption{Input-output format for GPT-35-Turbo.
This example is from the HellaSwag dataset.
We add some line breaks for better readability.}
\label{tab:app_gpt35_input}
\end{table*}

The hyperparameters for the retriever model and reward model are summarized in Table~\ref{tab:app_hyperparams}.
The E5$_\text{base}$ checkpoint is available at \url{https://huggingface.co/intfloat/e5-base-v2}.
This checkpoint is also employed for the k-means clustering baseline,
where we select $8$ examples closest to each cluster center as the in-context examples.
For each iteration,
we employ LLaMA-7B to rank the top-100 retrieved candidates.
As we retrieve from a unified pool of examples,
it is possible that a candidate comes from a different task than the query.
In this case,
we assign a low score to it.

During the evaluation,
we retrieve top-8 candidates and use them as in-context examples.
The maximum input length for LLaMA-7B is set to $1024$.
Longer inputs are truncated from the left side.
The maximum output length is set to $64$.
The most time-consuming part in our pipeline is
ranking candidates with LLaMA-7B,
which takes about $12$ hours for $200$k examples with 8 V100 GPUs.
Training the retriever model and reward model takes less than $10$ hours in total.

\section{Evaluation with GPT-35-Turbo} \label{sec:app_gpt35turbo}
Due to quota limits,
we sample at most $1$k examples for each dataset.
As GPT-35-Turbo does not return token-level log-probabilities,
we cannot evaluate the multiple-choice datasets
by computing the log-likelihood of each option.
Instead,
we append all the options to the end of the input,
and let the model generate the option index.
An example is shown in Table ~\ref{tab:app_gpt35_input}.
We also tried using this format to LLaMA-7B,
but the performance is significantly worse than comparing the log-likelihood of each option.

For a small number of test examples,
GPT-35-Turbo fails to follow the patterns of in-context examples
and generates outputs that are not valid class labels.
We add some simple heuristics based on string matching to determine the model prediction.

\begin{table*}[ht]
\centering
\scalebox{0.85}{\begin{tabular}{lcccccccccc}
\hline
\multirow{2}{*}{Task} & \multirow{2}{*}{Zero-shot} & \multirow{2}{*}{Random} & \multirow{2}{*}{Kmeans} & \multirow{2}{*}{BM25} & \multirow{2}{*}{E5$_\text{base}$} & \multirow{2}{*}{SBERT} & \multirow{2}{*}{EPR} & \multicolumn{3}{c}{LLM-R} \\ \cline{9-11}
 &  &  & &  &  &  & & 1 iter & 2 iter & 3 iter \\ \hline
AESLC & 5.8 & 19.4 & 19.0 & 26.8 & 27.0 & 25.3 & 26.0 & 26.7 & 27.3 & 27.1 \\
AGNews & 31.5 & 67.4 & 71.9 & 90.6 & 90.6 & 90.2 & 91.8 & 92.4 & 93.5 & 93.5 \\
ARC Chall. & 35.6 & 39.7 & 40.5 & 40.3 & 44.6 & 42.8 & 43.0 & 43.4 & 43.6 & 44.0 \\
ARC Easy & 51.0 & 60.0 & 61.8 & 59.9 & 63.0 & 63.1 & 63.1 & 63.6 & 63.3 & 63.6 \\
BoolQ & 64.7 & 70.0 & 69.0 & 74.7 & 72.4 & 73.9 & 74.8 & 75.6 & 75.1 & 74.1 \\
CommonGen & 19.2 & 36.3 & 34.4 & 37.6 & 37.4 & 37.6 & 39.2 & 38.2 & 37.7 & 37.3 \\
COPA & 66.0 & 80.0 & 85.0 & 78.0 & 83.0 & 82.0 & 82.0 & 84.0 & 84.0 & 84.0 \\
DART & 22.9 & 52.0 & 46.6 & 55.9 & 54.7 & 54.4 & 56.2 & 57.3 & 57.2 & 57.3 \\
E2E NLG & 34.6 & 52.7 & 46.4 & 54.5 & 51.8 & 50.2 & 53.6 & 54.9 & 54.7 & 54.9 \\
Gigaword & 15.3 & 30.0 & 30.7 & 32.7 & 32.5 & 32.6 & 32.4 & 33.3 & 32.5 & 31.8 \\
HellaSwag & 71.5 & 73.9 & 74.0 & 74.9 & 75.2 & 75.3 & 75.2 & 75.4 & 75.5 & 75.4 \\
MNLI (m) & 35.8 & 46.3 & 44.2 & 50.1 & 44.5 & 50.8 & 59.9 & 68.2 & 70.2 & 69.8 \\
MNLI (mm) & 35.6 & 48.1 & 45.4 & 48.3 & 44.7 & 49.3 & 61.5 & 69.5 & 72.0 & 71.3 \\
MRPC & 69.1 & 49.5 & 38.0 & 61.8 & 41.2 & 52.7 & 55.9 & 62.3 & 75.3 & 78.2 \\
MultiRC & 57.0 & 48.5 & 34.1 & 54.2 & 56.0 & 55.3 & 50.4 & 52.9 & 51.5 & 52.1 \\
NQ & 0.3 & 21.5 & 22.6 & 37.6 & 39.3 & 39.4 & 39.2 & 39.4 & 39.1 & 39.2 \\
OpenBook QA & 41.6 & 49.8 & 49.0 & 49.6 & 51.4 & 51.4 & 49.6 & 50.8 & 52.2 & 53.4 \\
PAWS & 53.2 & 57.0 & 56.6 & 56.6 & 55.4 & 58.2 & 57.7 & 57.0 & 56.6 & 57.0 \\
PIQA & 77.0 & 79.1 & 79.4 & 81.3 & 81.3 & 80.7 & 80.5 & 80.9 & 81.6 & 80.6 \\
QNLI & 49.2 & 56.4 & 53.4 & 62.2 & 61.5 & 61.9 & 65.0 & 74.4 & 69.6 & 69.4 \\
QQP & 57.7 & 63.4 & 63.3 & 79.8 & 77.5 & 81.3 & 81.7 & 80.1 & 82.6 & 83.3 \\
RTE & 59.6 & 59.9 & 58.5 & 65.7 & 63.9 & 67.2 & 66.8 & 67.2 & 68.6 & 70.4 \\
Sentiment140 & 49.3 & 88.6 & 89.4 & 90.8 & 93.9 & 92.2 & 91.4 & 90.8 & 91.1 & 90.3 \\
SNLI & 39.8 & 43.7 & 52.5 & 47.1 & 53.5 & 58.4 & 68.4 & 80.2 & 82.0 & 82.2 \\
SQuAD v1 & 2.1 & 64.1 & 62.3 & 61.2 & 60.8 & 61.6 & 64.3 & 60.7 & 57.3 & 52.5 \\
SST2 & 54.4 & 85.9 & 89.7 & 84.4 & 92.1 & 87.6 & 88.7 & 94.0 & 93.8 & 93.1 \\
Winogrande & 62.0 & 66.7 & 66.5 & 67.5 & 66.9 & 66.5 & 66.5 & 67.9 & 68.1 & 67.2 \\
WSC & 64.4 & 60.6 & 56.7 & 56.7 & 61.5 & 63.5 & 61.5 & 60.6 & 63.5 & 66.4 \\
WSC273 & 74.0 & 74.4 & 74.7 & 64.5 & 65.2 & 62.6 & 65.2 & 74.4 & 79.5 & 78.8 \\
Yelp & 47.9 & 92.0 & 93.5 & 93.5 & 97.3 & 95.9 & 95.1 & 95.7 & 95.9 & 95.5 \\ \hline
Average     &  44.9  & 57.9 & 57.0  & 61.3  & 61.4 &  62.1 & 63.5 & 65.7  &  66.5   & 66.4  \\ \hline
\end{tabular}}
\caption{Detailed results for each dataset.}
\label{tab:app_detailed_numbers}
\end{table*}

\begin{table*}[ht]
\centering
\scalebox{0.83}{\begin{tabular}{ll}
\hline
Task Name   & \begin{tabular}[c]{@{}p{1.0\linewidth}@{}} AG News \end{tabular} \\
Test Input  & \begin{tabular}[c]{@{}p{1.0\linewidth}@{}} "Holiday Shoppers Off to a Fast Start Holiday shoppers spent 10 percent more Friday than they did a year ago, according to early reports, but Wal-Mart Stores Inc. dampened hopes for a strong start to the key retail season by " What is this text about? World, Sports, Business, or Technology? \end{tabular} \\
Test Answer & \begin{tabular}[c]{@{}p{1.0\linewidth}@{}} \textbf{Business} \end{tabular} \\
LLM-R Top 1 & \begin{tabular}[c]{@{}p{1.0\linewidth}@{}} "Disappointing holiday news hurts retail shares Shares in a range of area retailers dipped Monday on disappointing Thanksgiving sales data from Wal-Mart Stores Inc. In addition, ShopperTrak, which tallies sales results from 30,000 stores nationwide, said " What is this text about? World, Sports, Business, or Technology?  \textbf{Business} \end{tabular} \\ \hline
Task name  & \begin{tabular}[c]{@{}p{1.0\linewidth}@{}} ARC Challenge \end{tabular} \\
Test Input  & \begin{tabular}[c]{@{}p{1.0\linewidth}@{}} In the 17th century, to estimate the distance to other planets, scientists first used the technique of viewing the planet from two different locations on Earth's surface. Which characteristic of the planet were the scientists using to calculate the distance from Earth? \end{tabular} \\
Test Answer  & \begin{tabular}[c]{@{}p{1.0\linewidth}@{}} \textbf{location} \end{tabular} \\
LLM-R Top 1 & \begin{tabular}[c]{@{}p{1.0\linewidth}@{}} Which physical characteristic of Earth is similar to a physical characteristic of the Moon? \textbf{its mountain ranges} \end{tabular} \\ \hline
Task name  & \begin{tabular}[c]{@{}p{1.0\linewidth}@{}} ARC Easy \end{tabular} \\
Test Input  & \begin{tabular}[c]{@{}p{1.0\linewidth}@{}} What is the major cause of seasonal changes? \end{tabular} \\
Test Answer  & \begin{tabular}[c]{@{}p{1.0\linewidth}@{}} \textbf{tilt of the Earth's axis} \end{tabular} \\
LLM-R Top 1 & \begin{tabular}[c]{@{}p{1.0\linewidth}@{}} Which occurs as a result of Earth's tilt on its rotating axis? \textbf{seasonal changes in the climate} \end{tabular} \\ \hline
Task name  & \begin{tabular}[c]{@{}p{1.0\linewidth}@{}} CommonGen \end{tabular} \\
Test Input  & \begin{tabular}[c]{@{}p{1.0\linewidth}@{}} Concepts: field, throw, kid, bunch, ball. Write a sentence that includes all these words. \end{tabular} \\
Test Answer  & \begin{tabular}[c]{@{}p{1.0\linewidth}@{}} \textbf{A bunch of kids are running around and throwing a ball on a field.} \end{tabular} \\
LLM-R Top 1 & \begin{tabular}[c]{@{}p{1.0\linewidth}@{}} Concepts: look, ball, lot. Write a sentence that includes all these words. \textbf{Two babies look up while they are playing in a playpen with a lot of balls.} \end{tabular} \\ \hline
Task name  & \begin{tabular}[c]{@{}p{1.0\linewidth}@{}} COPA \end{tabular} \\
Test Input  & \begin{tabular}[c]{@{}p{1.0\linewidth}@{}} "The boy skipped dinner." What is the cause? \end{tabular} \\
Test Answer  & \begin{tabular}[c]{@{}p{1.0\linewidth}@{}} \textbf{He ate a big lunch.} \end{tabular} \\
LLM-R Top 1 & \begin{tabular}[c]{@{}p{1.0\linewidth}@{}} "The parents left their children with a babysitter." What is the cause? \textbf{They made plans to celebrate their anniversary.} \end{tabular} \\ \hline
Task name  & \begin{tabular}[c]{@{}p{1.0\linewidth}@{}} DART \end{tabular} \\
Test Input  & \begin{tabular}[c]{@{}p{1.0\linewidth}@{}} Triple: The Mill, eatType, coffee shop; The Mill, food, Chinese; The Mill, priceRange, moderate; The Mill, area, city centre; The Mill, near, The Sorrento What is a sentence that describes this triple? \end{tabular} \\
Test Answer  & \begin{tabular}[c]{@{}p{1.0\linewidth}@{}} \textbf{There is a coffee shop serving Chinese food called The Mill. It has a moderate price range is is find in the city centre near The Sorrento.} \end{tabular} \\
LLM-R Top 1 & \begin{tabular}[c]{@{}p{1.0\linewidth}@{}} Triple: The Mill, eatType, coffee shop; The Mill, food, Indian; The Mill, priceRange, cheap; The Mill, area, riverside; The Mill, near, The Sorrento What is a sentence that describes this triple? \textbf{The Mill coffee shop is located in the riverside area near The Sorrento. They serve Indian food at a cheap price.} \end{tabular} \\ \hline
Task name  & \begin{tabular}[c]{@{}p{1.0\linewidth}@{}} Gigaword \end{tabular} \\
Test Input  & \begin{tabular}[c]{@{}p{1.0\linewidth}@{}} Write a short summary for this text: the dollar and major european currencies traded within narrow ranges on tuesday on the london forex market , which was waiting for the easter holiday weekend and for us employment figures to be announced on friday , traders said in late afternoon . \end{tabular} \\
Test Answer  & \begin{tabular}[c]{@{}p{1.0\linewidth}@{}} \textbf{london forex market stable as market waits for easter us data} \end{tabular} \\
LLM-R Top 1 & \begin{tabular}[c]{@{}p{1.0\linewidth}@{}} Write a short summary for this text: the dollar was stable over-all early monday afternoon by comparison with morning levels on the london forex market , which was waiting for publication at the end of the week of us inflation figures , traders said . \textbf{dollar stable in london as market waits for us inflation data} \end{tabular} \\ \hline
Task name  & \begin{tabular}[c]{@{}p{1.0\linewidth}@{}} MRPC \end{tabular} \\
Test Input  & \begin{tabular}[c]{@{}p{1.0\linewidth}@{}} Here are two sentences: An episode is declared when the ozone reaches .20 parts per million parts of air for one hour . A Stage 1 episode is declared when ozone levels reach 0.20 parts per million . Do they have the same meaning? \end{tabular} \\
Test Answer  & \begin{tabular}[c]{@{}p{1.0\linewidth}@{}} \textbf{Yes} \end{tabular} \\
LLM-R Top 1 & \begin{tabular}[c]{@{}p{1.0\linewidth}@{}} Here are two sentences: A Stage One alert is declared when ozone readings exceed 0.20 parts per million during a one-hour period . A Stage 1 episode is declared when ozone levels reach 0.20 parts per million . Do they have the same meaning? \textbf{Yes} \end{tabular} \\ \hline
Task name  & \begin{tabular}[c]{@{}p{1.0\linewidth}@{}} NQ \end{tabular} \\
Test Input  & \begin{tabular}[c]{@{}p{1.0\linewidth}@{}} Question: legislation regarding data protection and security in uk? Answer: \end{tabular} \\
Test Answer  & \begin{tabular}[c]{@{}p{1.0\linewidth}@{}} \textbf{The Data Protection Act 1998} \end{tabular} \\
LLM-R Top 1 & \begin{tabular}[c]{@{}p{1.0\linewidth}@{}} Question: which law relates to the protection of personal information? Answer: \textbf{Data Protection Act 1998} \end{tabular} \\ \hline
\end{tabular}}
\caption{More retrieved examples.
The format is the same as Table ~\ref{tab:cases}.}
\label{tab:app_retrieved_examples}
\end{table*}

\end{document}